\documentclass[conference]{IEEEtran}
\IEEEoverridecommandlockouts
\usepackage{cite}
\usepackage{amsmath,amssymb,amsfonts}
\usepackage{algorithmic}
\usepackage{graphicx}
\usepackage{textcomp}
\usepackage{xcolor}
\usepackage{lipsum}
\usepackage{booktabs} 

\usepackage{hyperref}
\usepackage{url}
\usepackage{graphicx}
\usepackage{amsmath,bm,amsthm}
\usepackage{amsthm}

\usepackage{amsmath}
\usepackage{amssymb}	
\usepackage{siunitx}
\usepackage{amsmath,bm,amsthm}
\usepackage{amsfonts}       
\usepackage{nicefrac}       
\usepackage{tikz}
\usepackage{mathtools}
\usepackage{booktabs} 
\usepackage{multirow}

\newcommand{\mysetminus}{\hbox{\tikz{\draw[line width=0.2pt,line cap=round] (1.5pt,0) -- (0,3pt);}}}

\newcommand{\Real}{\mathbb R}

\newcommand{\too}{\rightarrow}

\def\BibTeX{{\rm B\kern-.05em{\sc i\kern-.025em b}\kern-.08em
    T\kern-.1667em\lower.7ex\hbox{E}\kern-.125emX}}
\begin{document}

\title{Molecule Property Prediction and Classification with Graph Hypernetworks}

\author{\IEEEauthorblockN{Eliya Nachmani}
\IEEEauthorblockA{\textit{Facebook AI Research and Tel Aviv University} \\
enk100@gmail.com}
\and
\IEEEauthorblockN{Lior Wolf}
\IEEEauthorblockA{
\textit{Facebook AI Research and Tel Aviv University} \\
liorwolf@gmail.com}
}

\maketitle

\begin{abstract}
Graph neural networks are currently leading the performance charts in learning-based molecule property prediction and classification. Computational chemistry has, therefore, become the a prominent testbed for generic graph neural networks, as well as for specialized message passing methods. In this work, we demonstrate that the replacement of the underlying networks with hypernetworks leads to a boost in performance, obtaining state of the art results in various benchmarks.

A major difficulty in the application of hypernetworks is their lack of stability. We tackle this by combining the current message and the first message. A recent work has tackled the training instability of hypernetworks in the context of error correcting codes, by replacing the activation function of the message passing network with a low-order Taylor approximation of it. We demonstrate that our generic solution can replace this domain-specific solution.
\end{abstract}


\section{Introduction}

The field of learning-based prediction of molecule properties holds the promise of delivering accurate predictions at a fraction of the complexity that is required by the Density Functional Theory (DFT) models, while not being tied to the assumptions and approximations of this theory. This order of magnitude reduction in runtime supports not only the rapid screening of molecule banks, with important applications in medicine, manufacturing, and environmental science, but also the automatic design of new materials.

Molecules are often represented as graphs. Similarly to other application fields, such as computer vision, computational chemistry has benefited both from the development of powerful generic (graph) neural networks, as well as the development of specialized methods that are developed for the specific prediction tasks. For example, the state of the art NMP-Edge method of~\cite{edges} is a sophisticated domain-specific method, with many significant algorithmic choices, which generalizes the method of~\cite{schutt2017schnet} and incorporates ideas from the work of~\cite{gilmer2017neural} and \cite{kearnes2016molecular}.

In this work, we propose a generic way to improve graph neural networks and demonstrate that it is able to improve molecule property prediction and classification in both specialized networks and in more generic methods that are applied to computational chemistry datasets. The value of our scheme stems from its ability to improve upon a diverse set of already optimized state of the art methods.

Our method employs hypernetworks, also known as dynamic networks. In such neural networks, the weights of at least some of the layers vary dynamically based on the input. A hypernetwork can be seen as a composite network in which one network predicts the weights of another network. In our case, both networks receive the messages that are passed in the graph neural network as inputs.

Since the weights of the message generating network in the hypernetworks change dynamically during inference, training it is a challenge. We tackle this with a specific way of incorporating the incoming messages into the hypernetwork. Instead of passing the current message, we pass a linear combination of the current message and the first message. This simple modification is enough to ensure an improvement in performance. Without it, the hypernetwork would not typically outperform the original network.

Our experiments show that the same scheme is able to improve the predictions provided by three state of the art methods: the NMP-Edge network, the Invariant Graph Network of~\cite{maron2019provably}, and the Graph Isomorphism Network of~\cite{xu2018how}. In addition, we evaluate our method in the domain of error correcting codes, in which a very recent contribution by~\cite{nachmani2019} employed hypernetworks to improve the accuracy of a message passing scheme. We are able to show that our modification of the input messages is able to replace the method used there to stabilize the network. Taking both methods together, the results further improve.

\section{Related Work}

\noindent{\bf Graph Networks\quad} The topic of graph neural networks has drawn considerable attention, both in the context of specific applications, such as text analysis~\cite{socher2013recursive} and computer vision~\cite{johnson2018image} and as a generic tool. Earlier work employed recursive neural networks~\cite{goller1996learning,gori2005new,scarselli2008graph,li2015gated}, where the information flows once in a network that is generated based on the graph. Most current methods, including graph spectra methods~\cite{bruna2013spectral,defferrard2016convolutional,kipf2016semi} can be cast as message passing algorithms~\cite{gilmer2017neural,duvenaud2015convolutional,li2015gated,battaglia2016interaction,kearnes2016molecular,schutt2017quantum,xu2018how}.

The generic message passing methods employ three networks: one that pools the hidden states from the neighborhood graph vertices, another that updates the hidden states based on the aggregated representation of the neighbouring vertices, and one that reads the information from the entire graph in order to generate the final classification. In such a network, the messages are the hidden states of the nodes. In molecule prediction, conditioning the messages also on the receiving graph node and storing a hidden state for the linking edge improves performance~\cite{kearnes2016molecular,gilmer2017neural,edges}. 

An alternative to message passing techniques, is presented by permutation equivariant operators on the tensors that represent $k$-order interactions between graph nodes~\cite{kondor2018covariant,maron2018invariant,murphy2019relational,maron2019provably}. We are able to demonstrate that applying our method to both message passing methods and equivariant operator methods improved performance.

\noindent{\bf Molecule property prediction\quad} While in the past, feature engineering was the main route to applying machine learning in chemistry ~\cite{rogers2010extended,rupp2012fast,montavon2012learning,hansen2015machine,huang2016communication}, neural networks have become increasingly popular. The NMP-Edge model of~\cite{edges} described in Sec.~\ref{sec:edges} is an example of a specialized model. It follows the basic architecture of~\cite{gilmer2017neural}, in which the messages are conditioned on the nodes across both sides of the edge and on the hidden representation of the edge. In the NMP-Edge model, however, similar to SchNet~\cite{schutt2017schnet}, the message is an elementwise product of a network that encodes the sending node and a network that encodes the edge (not encoding the receiving edge directly). Also similar to SchNet, an RBF initialization and a soft-plus activation are used. Unlike SchNet, the edge embedding is being updated in time, following the Weave network proposed by~\cite{kearnes2016molecular}.

An example of a generic graph network solution that also excels on the popular QM9 benchmark  \cite{ramakrishnan2014quantum,ruddigkeit2012enumeration} is the Invariant Graph Network (IGN) of~\cite{maron2019provably}, presented in Sec.~\ref{sec:ign}.

\noindent{\bf Hypernetworks \quad}
Dynamic layers, also known as gating layers, are layers in which the weights are determined by a separate neural network. Such networks were introduced by \cite{klein2015dynamic,7410424} for visual tasks that require an adaptation of the input image. More recently, the term {\em hypernetworks} was coined to refer to a composite neural network in which a network $f$ is trained to predict the weights $\theta_g$ of another network $g$. The shift from specific layers to entire networks was presented by~\cite{jia2016dynamic}, who employed hypernetworks for video frames and stereo views prediction. The usage of hypernetworks for recurrent neural networks was presented by~\cite{ha2016hypernetworks}.~\cite{krueger2017bayes} have presented a Bayesian formulation of hypernetworks, and such networks have become prominent in meta-learning following~\cite{bertinetto2016learning}, who studied transfer learning between multiple few-shot learning tasks. 

Since the weights of network $g$ are generated instantaneously by network $f$, \cite{brock2018smash} have used hypernetworks for searching over the space of possible network architectures. In this case, a lengthy backpropagation optimization is replaced by the feed forward prediction of network $f$. More related to our work is that of~\cite{zhang2018graph}, who use  hypernetworks on graphs, also in the domain of network architecture search. In this work, the weight generating network $f$ is a graph network that operates on the graph that captures the generated architecture.  

Another recent application of hypernetworks to graphs is the work of \cite{nachmani2019}, where an MLP generates the weights of a message passing network that decodes error correcting codes. This generalizes earlier attempts in the domain of network decoders including~\cite{nachmani2016learning,kim2018deepcode,gruber2017deep,teng2018polar,cammerer2017scaling,vasic2018learning} and is shown to improve performance. The input to both the weight generating network $f$ and the message generation network $g$ is the incoming message, where for the first network, the absolute value is used. It is shown that training hypernetworks suffers from severe initialization challenges and would often lead to the explosion of the weights. \cite{nachmani2019}, therefore, present a new activation function that is more stable than the $arctanh$ activation typically used in message passing decoders. In our work, we employ conventional activations, and do not employ the absolute value for molecule prediction. We demonstrate that a combination of the initial message (from the first iteration) with the last message is an effective way to stabalize the training of the graph hypernetwork and do not employ dedicated activation functions.

\section{Graph Hypernetworks}

We extend three leading architectures for graph neural networks that were either designed for the molecule inference task or shown to excel on it. In each case, we add a hypernetwork scheme in which the input is a linear combination of the first message passed in the network and the current message. As our experiments show, in all three cases, sizable gains in performance are obtained, in comparison to the underlying method. 
In addition, in order to compare ourselves with a recent hypernetwork message passing scheme, we modify the decoding method of~\cite{nachmani2019}.

\subsection{Extending the NMP-Edge network by~\cite{edges}}
\label{sec:edges}

In order to describe how hypernetworks are applied to the NMP-Edge network,  we rely on the original notation of~\cite{edges}. Let $h^{t}_{v}$ be the hidden state of a node associated with a specific atoms at iteration $t$, and $e^{t}_{vw}$ be the hidden state representation of an edge, which denotes either a chemical link between atom or spatial proximity. The  hidden states of the atoms $h^{0}_{v}$ are initialized using a look-up table and the hidden state of the edges $e^{0}_{vw}$ are initialized using an RBF function with multiple scales, following~\cite{edges,schutt2017schnet}. 

The message passing scheme of the original NMP-Edge network takes the form:
\begin{align}
	m_{v}^{t+1} &= \sum_{w \in N(v)} M_{t}(h_w^t, e_{vw}^t), \label{eq:message_func}\\
	h_v^{t+1} &= S_t \left( h_v^t, m_{v}^{t+1} \right),
	\label{eq:updates}
\end{align}
where $m_v^t$ are the messages aggregated at node $v$ at time $t$, and $M_t$, $S_t$ are the message and transition networks for iteration $t$. These networks are dynamic (vary between iterations) but are independent of the inputs. The earlier work by~\cite{schutt2017quantum} uses a similar set of networks which do not change between the iterations.

In our modified network, we replace the state transition function with the hypernetwork $f$ and $g$ as follows:
\begin{equation}
\label{eq:edge_theta}
    \theta^{t}_{g}=f\left ( c \cdot  h^{0}_{v}  + (1-c) \cdot h^{t}_{v} \right )
\end{equation}
\begin{equation}
\label{eq:edge_ht}
    h^{t+1}_{v} = h^{t}_{v} + g_{\theta^{t}_{g}}\left ( m^{t+1}_{v} \right )
\end{equation}
where $c$ is a learned damping factor, which is clipped to be in the range [0,1] and is initialized with a uniform distribution, and the weights of network $g$ are given by $\theta^{t}_{g}$. For $t=0$ , Eq.~\ref{eq:edge_theta} becomes $\theta^{0}_{g}=f\left (h^{0}_{v}\right )$. Note that $f$ is a fixed function. However, $g_{\theta^{t}_{g}}$ vary in time, since the set of weights $\theta^{t}_{g}$ change as the input to $f$ changes.

The readout function is the same as in~\cite{edges}, which is two layer neural network that pools from all of the network atoms. The network $M_t$, the readout network, and network $S_t$ of the original architecture employ a shifted-soft-plus network, following~\cite{schutt2017schnet}, while $f,g$ employ $tanh$. Bias terms are not used. The number of layers is two in both $g$ and $S_t$. $f$ has four layers (we believe that the architecture of $S_t$ is locally optimal and adding layers to it did not improve the accuracy in our experiments).

\subsection{Extending the Invariant Graph Network of~\cite{maron2019provably}}
\label{sec:ign}
We follow the original notations of IGN, in order to enable a quick reference to the original IGN work. The original model with $d$ blocks has the form $F = m \circ h \circ B_d \circ ... \circ B_2 \circ B_1$ where, $h$ is an invariant layer~\cite{maron2019universality}, $m$ is a MLP, and the blocks $B_i$ are defined as:
\begin{equation}
\label{eq:ign_yi}
    Y_i = B_i(X_i) = [m_3(X_i),m_1(X_i) \odot m_2(X_i)]
\end{equation}
where $X_i\in\Real^{n\times n\times a}$ denote the input tensor to the block, $\odot$ denotes element-wise multiplication, the square brackets denote concatenation along the last tensor dimension, and the three MLPs $m_1,m_2:\Real^{a}\too \Real^{b}$ and $m_3:\Real^{a}\too \Real^{b'}$ are applied to each of the $n\times n$ elements of the input tensor individually along the third tensor dimension. The dimension of the block's output $Y_i$ is, therefore, $n \times n \times (b'+b)$.

The modified IGN network has the form:
\begin{equation}
\label{eq:hmb}
    F = m \circ h \circ B_d \circ H_{d-1} \circ ... \circ B_2 \circ H_1 \circ B_1, 
\end{equation}

where $H_1,...,H_{d-1}$ are hyper blocks. Let $Y_i=B_i(X_i)$ be the input tensor to the hyper block $H_i$. The hyper block $H_i$ performs the following computation:
\begin{equation}
\label{eq:ign_theta}
    \theta^{i}_{g} = f\left(c \cdot Y_{0} + (1-c)\cdot Y_{i} \right)
\end{equation}
\begin{equation}
\label{eq:ign_ht}
    H_i(Y_i) = g_{\theta^{i}_{g}}\left ( c \cdot Y_{0} + (1-c)\cdot Y_{i} \right )
\end{equation}
where the damping factor $c$ is a learned parameter, initialized from the uniform [0,1] distribution and clipped to remain in this range. The input tensors for $f$ and $g$ are an aggregation of $Y_0$ and $Y_i$ with the damping factor $c$. Each layer in $g$ is applied to each feature of the input tensor independently along the third dimension. As before, $f$ and $g$ are neural networks with the $tanh$ activation with $f$ having four layers and $g$ two. 

We use the same suffix networks per benchmark as \cite{maron2019provably}.  For QM9 $h$ of Eq.~\ref{eq:hmb} is an invariant max pooling, which is  followed by a MLP $m$ with three layers. For the classification datasets, $h$ is an  invariant max pooling layer from every block $B_i$ output $H_i(Y_i)$ ($Y_i$ in the original work) followed by a single layer. These outputs are then summed to produce the network output.

\subsection{Extending the Graph Isomorphism Network of~\cite{xu2018how}}

We now turn to the notation used in~\cite{xu2018how} to introduce the modified GIN model. $G(V,E)$ is the graph with node feature vector $X_v$ for vertices $v \in V$ and edges  $E$. In the graph classification problem, one is given a set of graphs with matching labels $\left \{(G_1, y_1), (G_2, y_2), ..., (G_N,y_N) \right \}$. The GNN model calculates representation vectors $h_v$ for each node $v$ in an iterative manner. After convergence, a readout function calculates the global graph embedding $h_G$, from which the label is predicted $\bar y_G=M(h_G)$ for a MLP $M$. In GIN, the readout takes the form of a summation followed by concatenation:
\begin{equation}
    h_G=\left [\sum_{v \in G}h^{(1)}_v,\sum_{v \in G}h^{(2)}_v, \dots,\sum_{v \in G}h^{(K)}_v \right ]
\end{equation}
where $K$ is number of message passing iterations. The update function of the hidden node representation for iteration $k$ is given by a MLP that is specific to this iteration:
\begin{equation}
    h_v^{(k)} =   {\rm MLP}^{(k)}   \left(  \left( 1 + \epsilon^{(k)} \right) \cdot h_v^{(k-1)} +  \sum\nolimits_{u \in \mathcal{N}(v)} h_u^{(k-1)}\right)
    \label{GIN_update}
\end{equation}
where $\epsilon^{(k)}$ is either a learned parameter or a fixed scalar, depending on the experiment.

The modified GIN model we propose modifies the update step, without changing the final readout:
\begin{equation}
    \theta^{k}_{g} =  f\left ( c \cdot h^{0}_{v} + (1-c) \cdot \left ( h_v^{(k-1)} + \sum\nolimits_{u \in \mathcal{N}(v)} h_u^{(k-1)}\right )    \right )
\end{equation}
\begin{equation}
    h_v^{(k)} =  g_{\theta^{k}_{g}}\left ( c \cdot h^{0}_{v} + (1-c) \cdot \left ( h_v^{(k-1)} + \sum\nolimits_{u \in \mathcal{N}(v)} h_u^{(k-1)}\right ) \right )
\end{equation}
where $h^{0}_{v}$ is calculated from Eq.~\ref{GIN_update} with $k=0$ as  $h_v^{(0)} = {\rm MLP}^{(0)}   \left(  \left( 1 + \epsilon^{(0)} \right) \cdot x_v +  \sum\nolimits_{u \in \mathcal{N}(v)} x_u\right)$, where $x_v$ is the vector of input features of node $v$. $c$ is a damping factor that is learned during training. $f$ and $g$ are neural networks with three and two hidden layers respectively with the $tanh$ activation. Note that the network $g$ changes between iterations and across nodes, depending on the input to $f$. However, the entire hypernetwork is fixed between the iterations.

\subsection{Extending the Decoding Hypernetwork of~\cite{nachmani2019}}

Since~\cite{nachmani2019} have proposed to extend an existing network using a hypernetwork, we modify their work in order to compare our way of converting a graph network to a hypernetwork with theirs. Specifically, it is reported that hypernetworks cannot train for the task of decoding error correcting codes, unless a dedicated activation function is used, since any other activation function attempted in their experiments leads to a divergence of the weights.

\cite{nachmani2019} modify the belief propagation algorithm of \cite{nachmani2016learning}, which is given by:

\begin{equation}
    x^{j}_e = x^{j}_{(c,v)} = \left\{\begin{matrix}
\tanh \left(\frac{1}{2}\left(l_v + \sum_{e'\in N(v)\setminus \{e\}} w_{e'}x^{j-1}_{e'}\right)\right), & \\ 
\hspace{5cm} $j$ \textup{ is odd}\\ 
2arctanh \left( \prod_{e'\in N(c) \setminus \{e\}}{x^{j-1}_{e'}}\right), & \\ 
\hspace{5cm} $j$ \textup{ is even} 
\end{matrix}\right.
\label{eq:2016_eq}
\end{equation}

where $l_v$ is the log likelihood ratios of the input bits, $x^j_e$ is the computed edge message for the edge $e=(c,v)$ in a Tanner graph, which is a bidirectional graph that has variable nodes $v$ on one side and check nodes $c$ on the other. Let $H$ be the parity check matrix. Each variable node is indexed by an edge $e=(c,v)$ on the Tanner graph and $N(v)=\{(c,v) | H(c,v)=1\}$, i.e, the set of all edges in which $v$ participates. $x^{j-1}_e$ is the message from the previous iteration. 

The hypernetwork model of \cite{nachmani2019} has the following update equations for odd $j$:
\begin{equation}
    \theta_g^j = f(|x^{j-1}|,\theta_f)
\end{equation}
\begin{equation}
    x^{j}_e = x^{j}_{(c,v)} = g(l_v,x^{j-1}_{N(v,\mysetminus c)},\theta_g^j),
\end{equation}
where $N(v,\mysetminus c)$ is a vector that contains the elements of $x^{j}$ that correspond to the indices $N(v)\setminus \{(c,v)\}$. For even $j$ the update equation takes the form:
\begin{equation}
\label{eq:even}
    x^j_e=x^j_{(c,v)} = 2\sum_{m=0}^{q}\frac{1}{2m+1}\left ( \prod_{e'\in N(c) \setminus \{(c,v)\}}{x_{e'}^{j-1}} \right )^{2m+1}
\end{equation}
where $q$ is the degree of the Taylor approximation of $arctanh$.

We modified the model of ~\cite{nachmani2019} with the following update equations. For odd $j$:
\begin{equation}
    \theta_g^j = f(|c \cdot x^{0} + (1-c) \cdot x^{j-1}|,\theta_f)
\end{equation}
\begin{equation}
    x^{j}_e = x^{j}_{(c,v)} = g(l_v,c \cdot x^{0} + (1-c) \cdot x^{j-1}_{N(v,\mysetminus c)},\theta_g^j),
\end{equation}
where $x^{0}$ is the output of one iteration from Eq.~\ref{eq:2016_eq}, and $c$ is the damping factor which is learned during training. 

For an even $j$ we either use Eq.~\ref{eq:even} (Taylor approximated arctanh), or consider the conventional $arctanh$ activation, as in Eq.~\ref{eq:2016_eq}. The readout function is not modified.

\section{Experiments}

We evaluate our model on regression and classification for predicting molecule proprieties and for decoding linear block codes. For regression we use the Quantum Machines 9 (QM9) dataset~\cite{ramakrishnan2014quantum, ruddigkeit2012enumeration} and Open Quantum Materials Database (OQMD) \cite{saal2013materials, kirklin2015open}.  The QM9 dataset has $133,885$ molecules. Each molecule has $12$ properties for predicting. When comparing our results to NMP-Edge and IGN methods, we use the same train-validation-test split as \cite{edges} or \cite{maron2019provably}, respectively. While based on the same dataset, these two benchmarks cannot be directly compared for many of the properties. The OQMD dataset contain $435,582$ inorganic structures. We use the same train-validation-test split as \cite{edges}.

For classification we employ the four bioinformatics dataset of \cite{yanardag2015deep}, which contains protein structures or chemical compounds: MUTAG, PROTEINS, PTC and NCI1. We use the original train folds, which are also used by \cite{maron2019provably} and \cite{xu2018how}.   

For decoding error correcting codes, we use the parity check metrics of \cite{channelcodes}. We use three classes of linear block codes: Low Density Parity Check (LDPC) codes \cite{gallager1962low}, Polar codes  \cite{arikan2008channel} and Bose-Chaudhuri-Hocquenghem (BCH) codes \cite{bose1960class}.

We compare with various baseline method on top of the methods that we modify. For the QM9 and OQMD datasets we compare with V-RF \cite{ward2017including}, SchNet \cite{schutt2017schnet}, enn-s2s \cite{gilmer2017neural}, Cormorant \cite{anderson2019cormorant}, Incidence \cite{albooyeh2019incidence}, 123-gnn \cite{morris2019weisfeiler} and the two methods by~\cite{wu2018moleculenet}: DTNN and MPNN. The baseline method for the classification datasets are WL subtree \cite{shervashidze2011weisfeiler}, DCNN \cite{atwood2016diffusion}, PATCHY-SAN\cite{niepert2016learning}, DGCNN \cite{zhang2018end}, AWL \cite{pmlr-v80-ivanov18a}, GCN \cite{kipf2016semi}, and GraphSAGE \cite{hamilton2017inductive}.

\noindent{\bf Implementation details} The various hyperparameters were selected based on the validation set (where vary between experiments) or set arbitrarily based on the underlying architecture (where fixed). \textbf{NMP-edge network} For QM9, we trained the models with the ADAM optimizer, with a learning rate set to $1e-4$. The number of iterations was $4$. The number of neurons in  network $f$ was $64$ and the number of neurons in network $g$ was $128$. The learning rate decreases by a factor of $0.96$ every $100,000$ gradient steps. The minibatch size was $32$. The node embedding size was $256$ to all the parameters. For OQMD, we use an ADAM optimizer, with a learning rate set to $1e-4$. The learning rate decreases by a factor of  $0.96$ every $400,000$ gradient steps. We use a minibatch of $32$ examples. The number of iterations was $3$. The number of neurons per layer in network $f$ was $64$  and that number in network $g$ was $128$. \textbf{Invariant graph network} For QM9, we use the same configuration as \cite{maron2019provably}, except for the following hyper parameters.  When training one model to predict all molecule parameters $f$ has four layers with $128$ neurons, whereas $g$ has two layers with $128$ neurons. When we trained a separate model for each molecule parameter, which calls for a smaller capacity, $f$ has four layers with $64$ neurons and $g$ has two layers with $64$ neurons. 
For the classification datasets we trained the models with the following hyperparameters, learning rate was $5e-5$ for MUTAG, PTC and NCI1, and was $1e-3$ for PROTEINS. The number of channels in blocks $B_i$ was $400$ for all datasets except for PROTEINS which has $128$ channels. The number of layers in the MLP $m$ was $2$ for all datasets. For all datasets the number of neurons in network $f$ was $64$ and the number of neurons in network $g$ was $64$, except for PROTEINS which has $32$ neurons for $f$ and $g$. \textbf{Graph isomorphism network}
For the classification datasets, we use the same training procedure as \cite{xu2018how}, who train the model for 10 folds and choose the number of epochs based on the cross validation accuracy over the folds. 
We train the models with the following hyper-parameters. Learning rate was $5e-3$, the models run for $5$ iterations and the number of epochs was $180$ for all datasets. The minibatch sizes were $512$, $256$, $32$ and $16$ for NCI1, PTC, MUTAG and PROTEINS, respectively. In all datasets, $f$ has three layers and $64$ neurons, $g$ has two layers with $64$ neurons each. \textbf{Decoding hyper-network} We use the same hyper-parameters as in \cite{nachmani2019}.

\subsection{Results}

\paragraph{Graph regression} The results for the QM9 dataset are reported in Tab.~\ref{tab:qm9_edge} for the NMP-Edge compatible splits and units and in Tab.~\ref{tab:qm9_ign} for the benchmark version used by IGN. Our model based on the NMP-Edge architecture achieves state of the art performance on 9 out of 12 parameters, and in only one parameter it is outperformed by the original NMP-Edge model (and another tie). 

The result for formation energy predictions OQMD, based on the NMP-Edges architecture, is provided in Tab.~\ref{tab:oqmd}. We obtain state of the art performance in this benchmark as well.

The QM9 model that is based on IGN obtains state of the art performance on 7 out of 12 parameters when training the model for each parameter. Furthermore, we improve the results of 9 out of 12 parameters when comparing to IGN model that is trained for each parameter separately. When training one model to predict all the parameters, we improve 12 out of 12 parameters, compared to the IGN model.

\paragraph{Graph classification} The results for the classification datasets are provided in Tab.~\ref{tab:classification_table}. As can be observed, our modified versions of IGN and GIN improve the baseline IGN and GIN models in almost all cases (in one case we tie). Note that the GIN model has many variants and our modification is based on the GIN-$\epsilon$ model. There is no Graph Neural Network model that outperforms our results, and for the PTC and PROTEINS datasets, our method outperforms all literature baselines. 

\begin{figure*}[t]
\centering
\begin{center}$
\begin{array}{ccc}
\hspace*{-1cm}
\includegraphics[width=.35\textwidth]{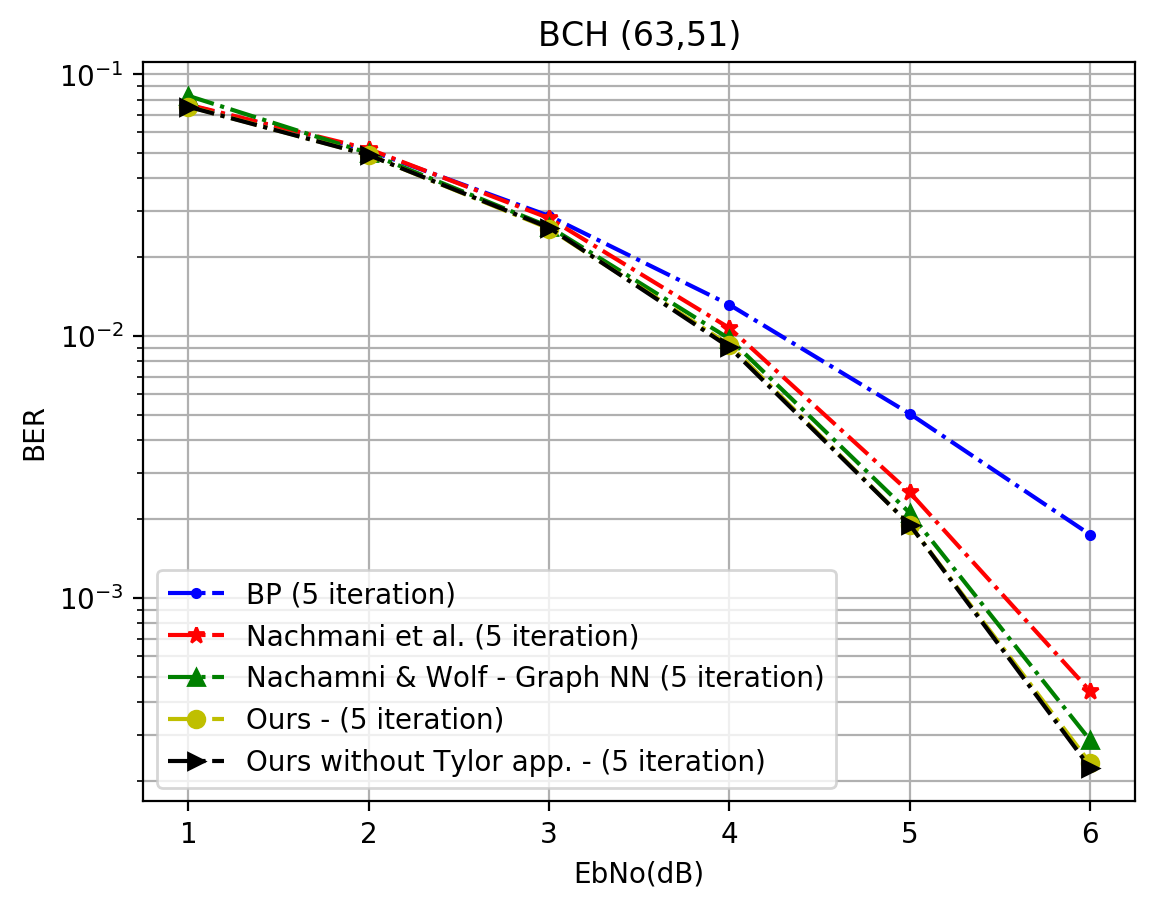} &
\includegraphics[width=.35\textwidth]{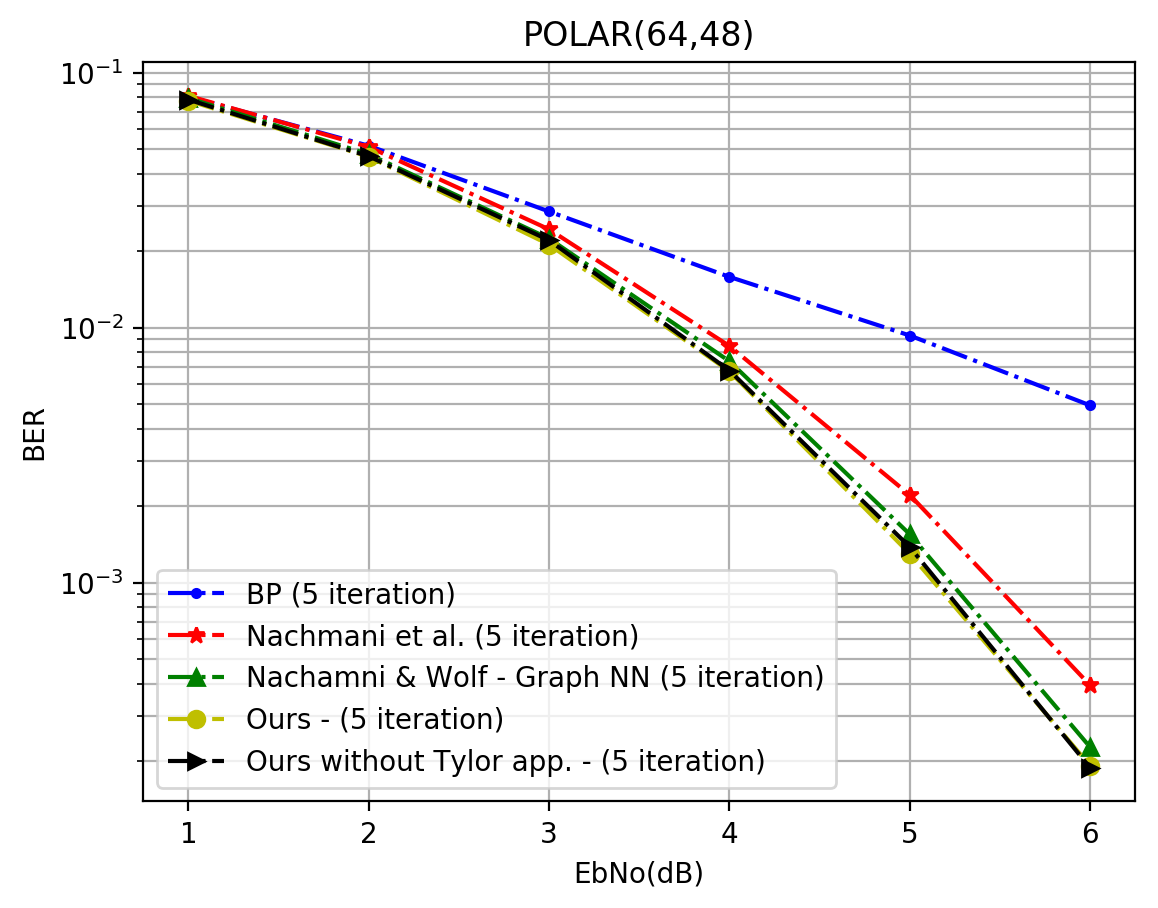} & 
\includegraphics[width=.35\textwidth]{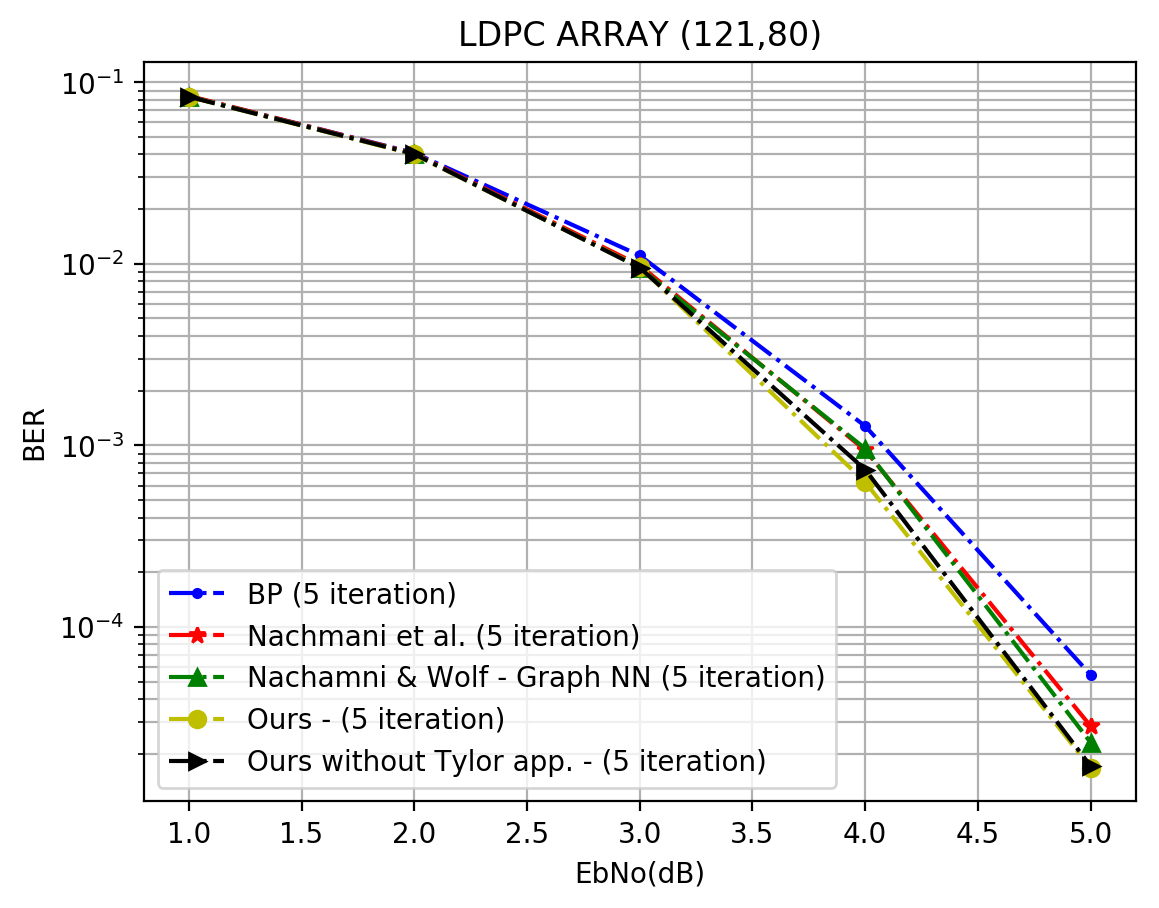} \\
(a) & (b) & (c) \\
\end{array}$
\end{center}
\caption{BER for various values of SNR for various codes. (a) BCH (63,51), (b) POLAR(64,48), (c) LDPC ARRAY (121,80).}
\label{fig:ber_snr}
\end{figure*}

\paragraph{Error Correcting Codes results} In Fig.~\ref{fig:ber_snr} we provide the BER-SNR results for multiple linear block codes. Our method improves on \cite{nachmani2019} across codes. We get an improvement range between $0.09dB$ and $0.12dB$ for large SNR. Moreover, in all three cases, we are able to improve the baseline results, even without the Taylor approximation of \cite{nachmani2019}. Since \cite{nachmani2019} fail to train without the approximation (using $arctanh$ their runs always diverge), this shows that our method stabilizes the training process for error correcting codes. We can also observe that our results with and without this approximation are almost identical.

\subsection{Ablation Analysis}
In Tab.~\ref{tab:qm9_ablation} we provide an ablation analysis on the QM9 benchmark. 
For the NMP-Edge model, we can observe degradation of 11 out of 12 parameters when training without $h^0_v$ in Eq.~\ref{eq:edge_theta} and the associated damping factor. Moreover, when training without the hypernetwork, but with $h^0_v$ (Eq.~\ref{eq:updates} becomes $ h_v^{t+1} = c \cdot h_v^{0} + (1-c) \cdot S_t \left( h_v^t, m_{v}^{t+1} \right) $) we get a degradation of 7 out of 12 parameters.  

For the IGN model, we can observe degradation in 9 out of the 12 parameters when training without the damping factor and $Y_0$ in Eq.~\ref{eq:ign_theta},~\ref{eq:ign_ht}. Moreover, when training without the hypernetwork but with the added first message ($X_i$ become $c \cdot X_0 + (1-c) \cdot X_i$ in Eq.~\ref{eq:ign_yi}), we get a degradation of 8 out of the 12 parameters.

\section{Conclusions}

Graph neural networks are becoming the dominant tool in molecule prediction and classification tasks. Here we show that by employing hypernetworks with a stabilization mechanism, significant performance gains are obtained. In order to demonstrate the advantage of our stabilizing mechanism over a recently proposed hypernetwork scheme, we also show improved performance in the field of decoding linear block codes.

\bibliographystyle{./bibliography/IEEEtran}
\bibliography{./bibliography/IEEEexample}

\clearpage
\newpage

\begin{table*}[t]
	\centering
	\caption{Mean absolute error for QM9 molecule parameters prediction, using the NMP-Edge splits, units and our modified architecture. The lowest error is in bold. The results of the Incidence and Cormorant networks are from \cite{albooyeh2019incidence} and \cite{anderson2019cormorant}, respectively. The rest of the results from \cite{edges}.}
	\label{tab:qm9_edge}
	\begin{tabular}{llllllll}
		\toprule
		Target & Unit & SchNet & enn-s2s & NMP-Edge & Cormorant & Incidence & Ours \\
		\midrule
		$\varepsilon_{HOMO}$  	& \si{meV}			& 41	& 43	& 36.7 & 36 & 89 & \textbf{26.56} \\
		$\varepsilon_{LUMO}$	& \si{meV}			& 34	& 37	& 30.8 & 36 & 49 & \textbf{23.47} \\
		$\Delta \varepsilon$	& \si{meV}			& 63	& 69	& 58.0 & 60 & 68 & \textbf{41.91} \\
		ZPVE					& \si{meV}			& 1.7	& 1.5	& \textbf{1.49} & 1.982 & 8 & \textbf{1.49} \\
		$\mu$					& \si{Debye}		& 0.033	& 0.030	& \textbf{0.029} & 0.130 & 0.04 & 0.031 \\
		$\alpha$				& \si{Bohr^3}		& 0.235	& 0.092	& 0.077& 0.092 & \textbf{0.03} & 0.066 \\
		$\langle R^2 \rangle$	& \si{Bohr^2}		& 0.073	& 0.180	& 0.072 &  0.673 & \textbf{0.017} & 0.057 \\
		$U_0$					& \si{meV}			& 14	& 19	& 10.5 & 28 & 8 & \textbf{7.27} \\
		$U$						& \si{meV}			& 19	& 19	& 10.6 & - & 7 & \textbf{6.99} \\
		$H$						& \si{meV}			& 14	& 17	& 11.3 & - & 8 & \textbf{7.17} \\
		$G$						& \si{meV}			& 14	& 19	& 12.2 & - & 8 & \textbf{7.99} \\
		$C_v$					& \si{cal\per molK}	& 0.033	& 0.040	& 0.032 & 0.031 & 0.028 & \textbf{0.026} \\
		\bottomrule
	\end{tabular}
\end{table*}

\begin{table*}
	\centering
	\caption{OQMD - NMP-Edge. Mean absolute error for formation energy predictions. The results of the various baselines are from \cite{edges}}
	\label{tab:oqmd}
	\begin{tabular}{llllll}
		\toprule
		Method: & V-RF & SchNet & NMP-Edge & Ours\\
		\midrule
		\si{meV \per atom}   & 74.5  & 27.5 &   14.9 &   \textbf{13.5}    \\
		\bottomrule
	\end{tabular}
\end{table*}

\begin{table*}
 \centering
  \caption{Mean absolute error for QM9 molecule parameters prediction for the IGN splits, units, and our modified architecture of it. The results of the various datasets are taken from \cite{maron2019provably}. For IGN and our modifications, we show results of a single network predicting all values and of dedicated networks.}
  \label{tab:qm9_ign}
  \centering
   \begin{tabular}{lrrrrrrr}
    \toprule
    Target & \multicolumn{1}{l}{DTNN} & \multicolumn{1}{l}{MPNN} & \multicolumn{1}{l}{123-gnn} & \multicolumn{1}{l}{IGN - all} & \multicolumn{1}{l}{IGN - single} & \multicolumn{1}{l}{Ours - all} & \multicolumn{1}{l}{Ours - single} \\
    \midrule
    $\mu$    & 0.244 & 0.358 & 0.476 & 0.231 & 0.0934 & \bf{0.157} & \bf{0.0883} \\
    $\alpha$ & 0.95  & 0.89  & \bf{0.27}  & 0.382 & 0.318 & 0.325 & 0.303 \\
    $\epsilon_{homo}$ & 0.00388 & 0.00541 & 0.00337 & 0.00276 & \bf{0.00174} & 0.00203 & 0.00178 \\
    $\epsilon_{lumo} $& 0.00512 & 0.00623 & 0.00351 & 0.00287 & 0.0021 & 0.00228 & \bf{0.0020} \\
    $\Delta_\epsilon $& 0.0112 & 0.0066 & 0.0048 & 0.00406 & 0.0029 & 0.00306 & \bf{0.0027} \\
    $\langle R^2 \rangle$ & 17    & 28.5  & 22.9  & 16.07 &  \bf{3.78} & 13.9 & 7.56 \\
    $ZPVE  $& 0.00172 & 0.00216 & \bf{0.00019} & 0.00064 & 0.000399 & 0.00049 & 0.000396 \\
    $U_0    $& 2.43  & 2.05  & 0.0427 & 0.234 & 0.022 & 0.093 & \bf{0.018} \\
    $U    $& 2.43  & 2     & 0.111 & 0.234 & 0.0504 & 0.092 & \bf{0.0174} \\
    $H     $& 2.43  & 2.02  & 0.0419 & 0.229 & 0.0294 & 0.093 & \bf{0.0193} \\
    $G     $& 2.43  & 2.02  & 0.0469 & 0.238 & 0.024 & 0.093 & \bf{0.017} \\
    $C_v  $& 0.27  & 0.42  & \bf{0.0944} & 0.184 & 0.144 & 0.180 & 0.146 \\
    \bottomrule
    \end{tabular}%
  \label{tab:reg_res}%
\end{table*}

\begin{table*}[t]
 \caption{Test set classification accuracy (\%) for MUTAG, PROTEINS, PTC, NCI1 datasets. The highest results on bold. The results of IGN are by \cite{maron2019provably}, the rest of the results are from \cite{xu2018how}. }
  \label{tab:classification_table}


\centering
\begin{tabular}{@{}clccccc@{}}\cmidrule[\heavyrulewidth]{2-7}

& Datasets & {\textsc{MUTAG}} &  {\textsc{PROTEINS}}  & {\textsc{PTC}} & {\textsc{NCI1}}  \\

\multirow{3}{*}{\rotatebox{90}{\hspace*{+3pt}Datasets}}
& \text{\# graphs }                      &  188                         & 1113          & 344           &  4110     \\
& \text{\# classes }                     &  2                           & 2             &  2            & 2         \\
& \text{Avg \# nodes }                   &  17.9                        & 39.1          & 25.5          &  29.8     \\
\cmidrule{2-7}
\multirow{5}{*}{\rotatebox{90}{\hspace*{-10pt}Baselines}}        
& \text{WL subtree}                      &  90.4 $\pm$ 5.7                & 75.0 $\pm$ 3.1   & 59.9 $\pm$ 4.3     & {\bf 86.0 $\pm$ 1.8} $^\ast$    \\
& \textsc{DCNN}                          &  67.0                          & 61.3             & 56.6               & 62.6                            \\
& \textsc{PatchySan}                     &  {\bf 92.6 $\pm$ 4.2} $^\ast$  & 75.9 $\pm$ 2.8   & 60.0 $\pm$ 4.8     & 78.6 $\pm$ 1.9                  \\ 
& \textsc{DGCNN}                         & 85.8                           & 75.5             & 58.6               & 74.4                            \\ 
& \textsc{AWL}                           & 87.9 $\pm$ 9.8                 & --               & --                 & --                              \\ 
\cmidrule{2-7}
\multirow{9}{*}{\rotatebox{90}{\hspace*{-10pt}GNN variants}}
& \textsc{Ours based on GIN}             & {\bf 90.55 $\pm$ 5.4}     & {\bf 76.90 $\pm$ 2.24}    & {\bf 69.68 $\pm$ 5.5}  & {\bf 82.7 $\pm$ 2.0}  \\
& \textsc{GIN-0}                         & {89.4 $\pm$ 5.6}          & {76.2 $\pm$ 2.8}          & {64.6 $\pm$ 7.0}       & {\bf 82.7 $\pm$ 1.7}        \\
& \textsc{GIN-$\epsilon$}                & {89.0 $\pm$ 6.0}          & {75.9 $\pm$ 3.8}          & 63.7 $\pm$ 8.2         & {82.7 $\pm$ 1.6}        \\
& \textsc{GIN-Sum--1-Layer}              &  {90.0 $\pm$ 8.8}         & {76.2 $\pm$ 2.6}          & 63.1 $\pm$ 5.7         & 82.0 $\pm$ 1.5          \\
& \textsc{GCN-Mean--1-Layer}             &  85.6 $\pm$ 5.8           & 76.0 $\pm$ 3.2            & 64.2 $\pm$ 4.3         & 80.2 $\pm$ 2.0   &      \\   
& \textsc{GraphSAGE-Max--1-Layer}        &  85.1 $\pm$ 7.6           & 75.9 $\pm$ 3.2            & 63.9 $\pm$ 7.7         & 77.7 $\pm$ 1.5   &      \\         
\cmidrule{2-7}
& \textsc{Ours based on IGN}    & {\bf 91.66 $\pm$ 6.54}    & {\bf 77.8 $\pm$ 5.93}          & {\bf 68.23 $\pm$ 10.07} & {81.99 $\pm$ 2.08}        \\
& \textsc{IGN}                  & {90.55 $\pm$ 8.7}         & {77.2 $\pm$ 4.73}         & {66.17 $\pm$ 6.54}     & {\bf 83.19 $\pm$ 1.11}      \\

\cmidrule[\heavyrulewidth]{2-7}
\end{tabular}
\end{table*}

\begin{table*}
	\centering
	\caption{Ablation analysis on the two QM9 dataset views. Mean absolute error is reported in both. }
	\label{tab:qm9_ablation}
	\begin{tabular}{@{}l@{\quad\quad}llll@{\quad\quad\quad}lll@{}}
		\toprule
		&\multicolumn{3}{c}{NMP-Edge split,units, and architecture} &\multicolumn{3}{c}{IGN split, units, and architecture}\\
		\cmidrule(lr){2-4}
		\cmidrule(lr){5-7}
		Target & Full & No $h_v^0$ & No hypernetwork & Full & No $X_0$ & No hypernetwork\\
		\midrule
		$\mu$               	& 0.031    & 0.033    & {\bf0.024}  & {\bf 0.157}   & 0.163 & 0.171 \\
		$\alpha$				& 0.066    & 0.455    & {\bf 0.065} & 0.325   & 0.332 & {\bf 0.318} \\
		$\varepsilon_{HOMO}$  	& 26.5     & 26.7     & {\bf 25.9}  & {\bf 0.00203} & 0.00223 & 0.00218 \\
		$\varepsilon_{LUMO}$	& 23.4     & 23.99    & {\bf 22.5}  & {\bf 0.00228} & 0.00229 &  0.00244 \\
		$\Delta \varepsilon$	& 41.916   & {\bf 34.44}    & 42.934 & {\bf 0.00306} & 0.00311 & 0.00341 \\
		$\langle R^2 \rangle$	& {\bf 0.057}    & 0.183    & 0.197 & 13.9    & {\bf 13.6} & 13.8 \\
		ZPVE					& {\bf 1.49}     & 2.16     & 1.51 & 0.00049 &  0.00045 & {\bf 0.00045} \\
		$U_0$					& {\bf 7.27}    & 7.71     & 9.09 & {\bf 0.093}   & 0.099 & 0.102 \\
		$U$						& {\bf 6.99}   & 7.73     & 9.64 & {\bf 0.092}   & 0.0986 & 0.103 \\
		$H$						& {\bf 7.17}    & 7.40     & 9.36 & {\bf 0.093}   & 0.096 & 0.104 \\
		$G$						& {\bf 7.99}     & 9.05     & 9.58 & {\bf 0.093}   & 0.0997 & 0.102 \\
		$C_v$					& 0.026    & 0.043    & {\bf 0.025} & 0.180   & {\bf 0.163} & 0.180 \\
		\bottomrule
	\end{tabular}
\end{table*}

\end{document}